\documentclass[twoside,11pt]{article}
\usepackage{subfigure}
\usepackage{array}
\usepackage{booktabs}
\usepackage{enumitem}
\usepackage{blindtext}
\usepackage[abbrvbib, preprint]{ArXiv}
\usepackage[svgnames]{xcolor}
\hypersetup{
hidelinks,
colorlinks=true,
linkcolor=Indigo,
urlcolor=DeepSkyBlue4,
citecolor=Indigo
}
\usepackage{todonotes}
\usepackage{amsmath} 

\usepackage{marvosym}
\makeatletter
\def\@fnsymbol#1{\ifcase#1\or \text{\Letter}\or *\or \dagger\or \ddagger\else\@arabic{#1}\fi}
\makeatother

\usepackage{lastpage}
\jmlrheading{23}{\ Preprint 2025}{1-\pageref{LastPage}}{1/21; Revised 5/22}{9/22}{21-0000}{}

\firstpageno{1}

\begin{document}

\title{Compressing Model with Few Class-Imbalance Samples: An Out-of-Distribution Expedition}

\author{\name Tian-Shuang Wu \email tianshuangwu@hhu.edu.cn \\
\addr Key Laboratory of Water Big Data Technology of Ministry of Water Resources,\\ College of Computer Science and Software Engineering, Hohai University, Nanjing, China\\
\AND
\name Shen-Huan Lyu~\thanks{Corresponding author} \email lvsh@hhu.edu.cn \\
\addr Key Laboratory of Water Big Data Technology of Ministry of Water Resources,\\ College of Computer Science and Software Engineering, Hohai University, Nanjing, China\\
\addr State Key Laboratory for Novel Software Technology,
Nanjing University, Nanjing, China\\
\AND
\name Ning Chen \email che-n-ing@hhu.edu.cn\\
\addr Key Laboratory of Water Big Data Technology of Ministry of Water Resources,\\ College of Computer Science and Software Engineering, Hohai University, Nanjing, China\\
\AND
\name Zhihao Qu \email quzhihao@hhu.edu.cn\\
\addr Key Laboratory of Water Big Data Technology of Ministry of Water Resources,\\ College of Computer Science and Software Engineering, Hohai University, Nanjing, China\\
\AND
\name Baoliu Ye \email yebl@nju.edu.cn\\
\addr State Key Laboratory for Novel Software Technology,
Nanjing University, Nanjing, China
       }

\editor{My editor}

\maketitle

\begin{abstract}
In recent years, as a compromise between privacy and performance, few-sample model compression has been widely adopted to deal with limited data resulting from privacy and security concerns. However, when the number of available samples is extremely limited, class imbalance becomes a common and tricky problem. Achieving an equal number of samples across all classes is often costly and impractical in real-world applications, and previous studies on few-sample model compression have mostly ignored this significant issue. Our experiments comprehensively demonstrate that class imbalance negatively affects the overall performance of few-sample model compression methods. To address this problem, we propose a novel and adaptive framework named OOD-Enhanced Few-Sample Model Compression (OE-FSMC). This framework integrates easily accessible out-of-distribution (OOD) data into both the compression and fine-tuning processes, effectively rebalancing the training distribution. We also incorporate a joint distillation loss and a regularization term to reduce the risk of the model overfitting to the OOD data. Extensive experiments on multiple benchmark datasets show that our framework can be seamlessly incorporated into existing few-sample model compression methods, effectively mitigating the accuracy degradation caused by class imbalance.
\end{abstract}


\section{Introduction}
As deep learning technology has advanced, models have become larger and more complex, resulting in challenges related to computational resources and storage in real-world applications. For instance, CNNs with millions of parameters cannot be deployed directly on edge devices like watches or cameras. Model compression has emerged as a key focus for achieving efficient inference with limited storage space. To compress the model, network pruning motheds~\citep{hao2017pruning,yihuiChannel2017} try to remove less significant weights or neurons, while knowledge distillation methods~\citep{hinton2015distilling} let the compact model learn from soft labels of the pre-trained model, and quantization methods~\citep{nagel2019data,adrianafitnet2015} try to reduce the precision of model weights and activations.

These compression methods have been successful in reducing the size of the model. However, they most commonly rely on a large number of samples to ensure the stability and accuracy of the model's performance. However, in real-world applications, particularly in sensitive fields such as medicine and finance, these methods are often confronted by privacy and security concerns. In addressing this problem, few-sample model compression has emerged as a prominent strategy for balancing privacy and performance and has received considerable attention in recent years. 

\begin{figure}[t]
\centering
\includegraphics[width=\columnwidth]{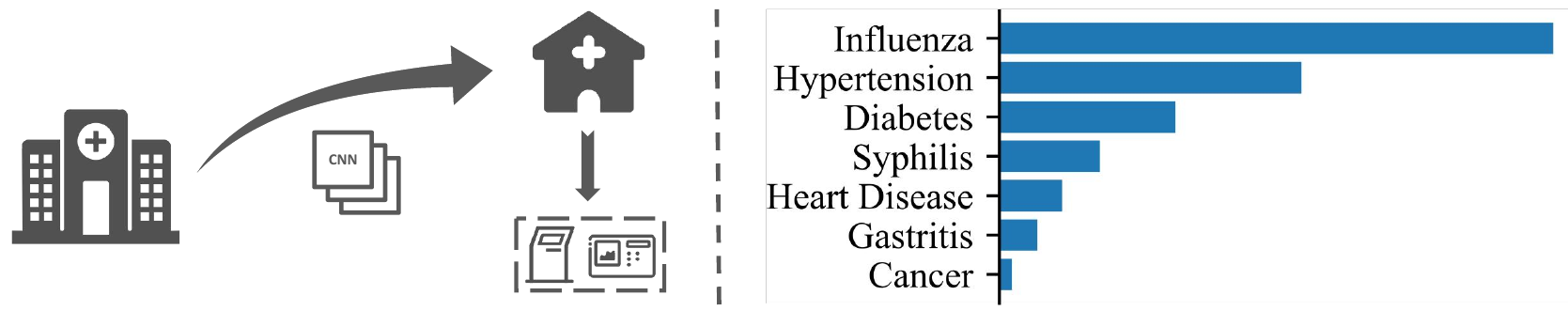}
\caption{Overview of introduction. \textbf{Left:} Illustration of few-sample model compression workflow for deployment in small hospitals. \textbf{Right:} Class imbalance in disease cases.}
\label{fig: introduction}
\end{figure}

Consider a small hospital that is planning to implement an efficient intelligent diagnostic system. Due to its limited sample size, the hospital cannot train a high-quality model independently. Moreover, it is prohibited to obtain complete case data from a larger hospital due to privacy policies. In such contexts, the deployment of few-sample compression methods has emerged as a promising solution. As shown in Figure~\ref{fig: introduction}, the large hospital can provide a pre-trained model based on its sufficient datasets, and then the small hospital can use its limited samples to compress and fine-tune this model, transforming it into a lightweight version. This kind of method facilitates efficient deployment of the model on intelligent diagnostic equipment.

While previous few-sample model compression strategies have shown promising results in addressing privacy concerns and optimizing model performance with limited data, they have not considered the problem of class imbalance in small sample settings. Existing research on few-sample model compression usually adopts the $N$-way $K$-shot experimental setup, where the training set contains $N$ classes, with $K$ samples assigned to each class. However, this setting overlooks the common issue of class imbalance in few-sample scenarios, where equal sample distribution across classes is often unrealistic in real-world applications. For instance, in the context of medical diagnosis, Figure~\ref{fig: introduction} illustrates a substantial imbalance in the number of cases. In the limited samples available, common diseases like the flu usually account for the majority, while rarer diseases such as cancer may have very few or even no samples at all. In this case, the problem caused by class imbalance is further amplified, which not only leads to biased training but also introduces bias into the compression process. Specifically, it causes the model to prioritize compressing the parts related to the minority class during the compression process.

In this paper, we first design experiments to demonstrate that class imbalance impairs few-sample compression. Subsequently, we propose a novel framework to address this problem by incorporating \emph{out-of-distribution} (OOD) data during the compression process to achieve dynamic balance, which we call OOD-Enhanced Few-Sample Model Compression (OE-FSMC). For each OOD instance, we sample labels from a predefined distribution that complements the original class priors. To mitigate the overfitting issue that may arise in the minority classes after introducing OOD data, we introduce a joint distillation framework to balance the relative contribution of the losses from the original data and the auxiliary dataset and we used a class-dependent weight to provide stronger regularization for the minority classes. This approach allows us to utilize noisy labels from the OOD data to rebalance the class priors while ensuring that these labels remain non-harmful to the compression and fine-tuning process. Experimental results demonstrate the effectiveness and wide applicability of our method. Our main contributions are summarized as follows: 
\begin{itemize}
\item To the best of our knowledge, this study is the first to address and solve the class imbalance problem in the few-sample model compression. 
\item We propose a novel framework for few-sample model compression based on out-of-distribution data, OOD-Enhanced Few-Sample Model Compression (OE-FSMC), for the class imbalance problem. In addition, we introduce a joint distillation framework and a class-dependent weight factor to mitigate the overfitting issue on minority classes.
\item We conduct extensive empirical studies on three datasets with class imbalance, and the experimental results show that our method can effectively improve the performance of existing few-sample compression methods when faced with the class imbalance problem. Furthermore, it has no restriction on the model's structure.
\end{itemize}

\section{Related Work}
\subsection{Few-sample Model Compression}
Few-sample model compression aims to derive compact models from pre-trained overparameterized networks using few samples. The main challenge is that the student model tends to overfit due to the scarcity of training data, leading to high inference errors. These errors progressively escalate through layer-wise propagation and accumulation~\citep{dong2017learning}, severely compromising output reliability.

To mitigate this problem, existing approaches employ layer-wise optimization to enforce intermediate-layer consistency between the compressed and original models. For instance, \citet{bai2020few} developed cross distillation (CD), which interleaves the teacher and student hidden layers to suppress inter-layer error propagation. Similarly, FSKD~\citep{li2020few} introduces learnable 
$1 \times 1$ convolutions at student network blocks, optimizing auxiliary parameters to bridge block-level representation gaps with the teacher mode. Such layer-wise optimization methods are computationally inefficient and not immune to the risk of error accumulation inherent in their design. 

Therefore, new approaches have gradually abolished the layer-wise reconstruction framework. For instance, MiR~\citep{wang2022mir} aligned outputs at the penultimate layer of teacher and pruned student models, then substituted all layers except the head before the penultimate layer in the teacher model with the trained student model. Another work~\citep{wang2023practical} replaced traditional filter pruning with block dropping and introduced a latency-accuracy evaluation metric, emphasizing the recovery efficiency of the compressed model and advocating for compression strategies with a high acceleration ratio. 

However, these methods overlook sample distribution imbalances, which can result in the loss of important information from minority classes during compression. Moreover, fine-tuning may exacerbate this bias towards majority classes, reducing the overall accuracy and weakening the generalization ability of the compressed model.

\subsection{Class Imbalance Problem}
In classification tasks, class imbalance occurs when the sample sizes of different classes vary significantly~\citep{ochal2023few}, causing the model to focus on majority classes and neglect critical minority-class information. Current solutions for addressing class imbalance problems can be classified into three main categories: data-level methods, algorithm-level methods, and hybrid approaches that combine both.

Data-level methods achieve class balance by either removing majority-class samples~\citep{liu2008exploratory,lin2017clustering,mohammed2020machine} or generating additional minority-class samples~\citep{chawla2002smote,sharma2022review,abdi2015combat}. These methods are simple and effective, making them the most commonly employed strategies. However, they also have limitations. For instance, removing majority-class samples may result in the loss of valuable information, leading to underfitting, while generating additional minority-class samples may cause overfitting~\citep{zhou2020bbn}, thereby impacting the model's generalization performance. 

Algorithm-level methods~\citep{zhou2005training,fernandez2018cost,he2024multi} attempt to mitigate the preference for majority classes by modifying existing machine learning algorithms. However, these methods often require extensive domain knowledge and experimentation, which may not be ideal for few-shot scenarios. Hybrid methods typically combine data-level or algorithm-level strategies with ensemble learning~\citep{galar2011review,chawla2003smoteboost}, taking advantage of the strengths of both strategies. Nevertheless, such hybrid methods inherit the limitations of the data-level or algorithm-level approaches and usually involve substantial computational overhead, making them difficult to implement on lightweight devices.

Although the class imbalance problem has been extensively studied across various tasks, research on the generalization performance of these methods in the context of few-sample model compression remains limited.

\section{Preliminaries}

\textbf{Few-sample Model Compression} aims to get a compact model from the pre-trained redundant model with few samples for multi-class classification, where the input space is represented by $ \mathcal{X} \in \mathbb{R}^d $, and the label space $ \mathcal{Y} $ consists of $ \{1, \dots, K\} $. Let $ D_{\text{full}} = \{(x_i, y_i)\}_{i=1}^N \in \mathcal{X} \times \mathcal{Y} $ denote the full dataset containing $ N $ samples, and $ D_{\text{few}} = \{(x_i, y_i)\}_{i=1}^M \in \mathcal{X} \times \mathcal{Y} $ represent the training dataset, consisting of $ M $ samples, where $ M \ll N $. Let $ m_j $ denote the number of samples of class $ j $ in $D_{\text{few}}$ , such that $ M = \sum_{j=1}^K m_j $. $ M_t $ is a redundant model pre-trained on the full dataset $ D_{\text{full}} $. 
Current mainstream few-sample model compression methods typically consist of two stages: compression and fine-tuning. The following analysis will be conducted separately for these two stages. In the compression stage, most methods combine pruning and knowledge distillation as the primary compression strategies, although there are cases where only one of these techniques is used. For instance, the PRACTISE method only employs pruning. To provide a more thorough introduction to the algorithmic process of few-sample model compression, the discussion in this section will be based on the methods including both pruning and knowledge distillation.

Taking a representative few-sample model compression method,  cross distillation~\citep{bai2020few}, as an example, in the compression stage, they use pruning methods to remove redundant channels or residual blocks from the $ M_t $, which effectively reduces the model's parameter count. Then they distill the knowledge from the $ M_t $ to the pruned model and make every layer have the same outputs as the corresponding layer in the $ M_t $, ensuring that the performance of the pruned model on the classification task can approximate that of the teacher model. In the final fine-tuning stage, they retrain the model on the training set to recover accuracy and further enhance performance, yielding a lightweight model. Throughout the entire process, they can only use a few-sample dataset $ D_{\text{few}}$.

\section{Method}

 \subsection{The Urgent Need to Address Class Imbalance in Few-Sample Model Compression}
Few-shot learning is particularly susceptible to class imbalance, as it is often impractical to expect an equal amount of data for each class when data is limited. The detrimental effects of class imbalance are well-documented, yet its impact within the context of model compression remains insufficiently explored. In the following, I will analyze, from a theoretical perspective, the adverse effects of class imbalance on model compression. Existing few-sample model compression techniques typically integrate network pruning, knowledge distillation, and fine-tuning. Therefore, I will individually assess the influence of class imbalance on each of these three techniques.

\begin{figure*}[ht]
\centering
\includegraphics[width=\linewidth]{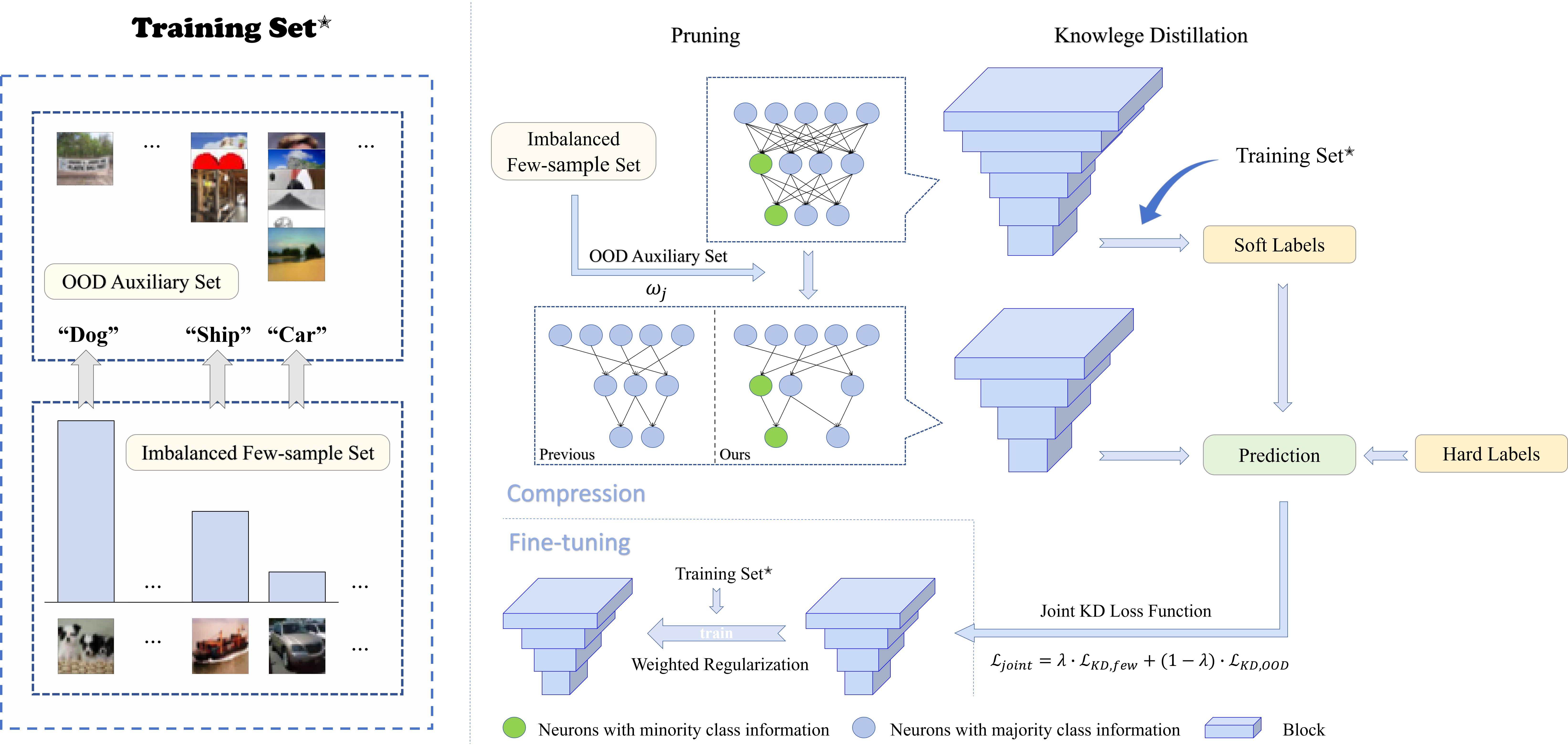}
\caption{Illustration of OE-FSMC. \textbf{Left:} Label assignment strategy for OOD data. \textbf{Right:} Framework of OE-FSMC in Compression and Fine-Tuning Process.} 
\label{fig:mainpic}
\end{figure*}
 
In network pruning, we compress the model by removing redundant filters or channels. Let the activation output of the $ l $-th layer be denoted as $ \mathbf{A}^{(l)} \in \mathbb{R}^{C_l \times H_l \times W_l} $, where $ C_l $ represents the number of channels (filters), and $ H_l $ and $ W_l $ denote the height and width of the feature map, respectively. The objective of pruning is to rank the filters based on some importance score $ s_k^{(l)} $ and remove the channels with the lowest scores. The importance score $ s_k^{(l)} $ is defined as:
\begin{equation}
\label{eq: pruning}
s_k^{(l)} = \sum_{j=1}^K \frac{m_j}{M} s_{k,j}^{(l)}\ , 
\end{equation}
where $ s_{k,j}^{(l)} $ represents the importance contribution of filter $ k $ to class $ j $. In the case of class imbalance, the term $ \frac{m_j}{M} $ is small for the minority class, which weakens its contribution. This leads to the erroneous pruning of filters associated with minority classes. The pruned filters are often essential for capturing features of the minority classes, resulting in a diminished feature representation for these classes in the pruned model, which further exacerbates the imbalance problem.

The objective of knowledge distillation is to minimize the discrepancy between probability distributions(soft labels) of the student and teacher models, typically measured using the Kullback-Leibler (KL) divergence:
\begin{equation}
\mathcal{L}_{\text{KD}} = \frac{1}{M} \sum_{i=1}^M \text{KL}\left(P_T(Y|x_i) \, \| \, P_S(Y|x_i)\right)\ ,  
\end{equation}
where $P_T(Y|x)$ represents the probability distribution output by the teacher model and $P_S(Y|x)$ represents the probability distribution output by the student model.

During the distillation process, the problem of class imbalance primarily manifests itself in the distribution of training data for the student model, while the soft labels of the teacher model are based on a balanced training set. Due to the pruning step, filters responsible for extracting features of the minority classes are removed, causing significant discrepancies between the soft labels of the student and teacher model for the minority classes. This leads to a costly and ineffective fitting process for the student model. Additionally, since the hard labels of the real-world data are imbalanced, the student model tends to focus more on learning knowledge related to the majority class during the distillation process. This is analogous to classroom learning, where students are more likely to focus on frequently appearing topics in exercises to ensure overall accuracy. However, this results in poorer performance on less frequent topics. Similarly, the student model often exhibits poor generalization performance on the minority classes.

In the fine-tuning process, the model tends to optimize for accuracy on the majority class while neglecting the minority classes, resulting in weaker prediction performance for the minority classes. In other words, the compressed model often has poor predictive performance on minority classes, and since only a very limited number of samples are available to restore accuracy, the fine-tuning process is almost ineffective in recovering the performance on minority classes.

Based on the above analysis, it is evident that the issue of class imbalance has a detrimental impact across various stages of few-sample compression, which cannot be overlooked. This imbalance leads to a degradation in the feature representation of minority classes, thereby negatively affecting the overall performance of the model. To address this, we propose an OOD-based framework aimed at mitigating the adverse effects of class imbalance on model compression techniques.

\subsection{Feasibility of Using OOD Data
}

Due to the lack of research on the class imbalance problem in few-sample model compression, we considered 
Class Imbalance Few Shot Learning (CIFSL). During this investigation, we discovered that certain resampling methods can generalize well to few-shot scenarios~\citep{ochal2023few}. Consequently, we examined popular resampling methods from recent years, such as ROS (Random Over-Sampling)~\citep{japkowicz2002class}, which duplicates samples from the minority class. However, these methods usually result in overfitting to minority classes. Other works have introduced synthetic samples to augment the minority class without repetition \citep{chawla2002smote} and some recent approaches have leveraged unlabeled data from the distribution to compensate for the lack of training samples, while these methods are highly costly in the context of few-sample model compression. Therefore, \citet{wei2022open} use OOD data to rebalance class priors and they rigorously proved that incorporating OOD data into the training set can be harmless through Theorem~\ref{theorem:1}:




\begin{theorem}
\label{theorem:1}
When labels are uniformly sampled from the label space within the distribution, augmenting the training set $D_{\mathrm{mix}} = D_{\mathrm{train}} \cup D_{\mathrm{out}}$ does not affect the prediction of the Bayesian classifier:
\begin{equation}
\mathop{\arg\max}_{y \in \mathcal{Y}} P_{\mathrm{mix}}(x|y) P_{\mathrm{mix}}(y) = \mathop{\arg\max}\limits_{y \in \mathcal{Y}} P_{\mathrm{s}}(x|y) P_{\mathrm{s}}(y)\ ,
\end{equation}
where $P_{\mathrm{mix}}(X, Y)$ represents the data distribution of $D_{\mathrm{mix}}$.    
\end{theorem}

Theorem~\ref{theorem:1} indicates that when labels are uniformly sampled from the label space of the in-distribution data, augmenting the training set with OOD instances does not alter the Bayesian classifier's prediction. Based on this, we can infer that the feature space of the student model remains aligned with that of the teacher model on $D_{\mathrm{mix}}$ after incorporating OOD instances during model compression. Therefore, we advocate the use of easily accessible out-of-distribution (OOD) data to rebalance the training data and mitigate the class imbalance problem in the few-sample model compression.

Inspired by~\citep{wei2022open}, we employ a complementary distribution as the label assignment strategy for OOD data. This strategy allocates more OOD instances to the minority class while ensuring the stability of the Bayesian classifier’s prediction. Specifically, the complementary sampling rate $\Gamma_j$ for class $j$ is defined as: 
\begin{equation}
\label{eq:labelassign}
\Gamma_j = \frac{\left( \alpha - \beta_j \right)}{\left( K \cdot \alpha - 1 \right)}\ ,
\end{equation}
where $\alpha = \max_j(\beta_j) + \min_j(\beta_j)$ represents a trade-off between the new balanced class prior and maintaining OOD data without degrading the model's performance, $\sum_{i=1}^K \Gamma_i = 1$ ensures that the sampling rates are normalized, and $\beta_j = \frac{m_j}{M}$ represents the original distribution weight for class $j$. 

This assignment strategy makes the class distribution of the mixed dataset $P_{\mathrm{mix}}(y)$ approximates the balanced distribution of the test set $P_{\mathrm{t}}(y)=\frac{1}{K}$, which ensures that all classes receive equal attention during the model compression process and alleviates the tendency to overlook minority classes. It is evident that the strategy is equally effective during the fine-tuning stage. 
These theoretical results provide a solid foundation for us to design a framework based on OOD data to tackle the class imbalance problem encountered during the few-sample model compression process.

\subsection{Our Method: A Framework Using OOD Data
}

In this section, we propose a novel framework called OOD-Enhanced Few-Sample Model Compression (OE-FSMC) for addressing the class imbalance in the few-sample model compression, leveraging the Open Sampling technique. As shown in Figure~\ref{fig:mainpic}, our approach consists of three key components: (1) \textbf{OOD Set Handling}, (2) \textbf{Compression Stage}, and (3) \textbf{Fine-tuning Stage}.

\paragraph{OOD Set Handling:} Previous studies commonly utilized the entire TinyImage dataset, which contains approximately 300K randomly sampled images, as the OOD dataset. However, under the constraints of model compression, limited storage capacity makes it impractical to store such a large dataset. To address this, we pre-sample 500 images from the OOD dataset using random sampling. During this process, we ensure the purity of the OOD dataset by removing any overlapping samples with the full training dataset $ D_{\text{full}} $. The experimental section provides a detailed analysis of the impact of $ n $ (the size of the OOD dataset) and demonstrates that reducing the OOD dataset size has a negligible effect on accuracy.

\paragraph{Compression stage:} When compressing the pre-trained teacher model, we randomly sample $ m $ (discussed further in the experimental section) instances from the OOD dataset to construct an auxiliary dataset. Labels for these OOD samples are assigned based on Equation~\ref{eq:labelassign}, ensuring balanced attention across all classes during the compression process.

During network pruning, the sparse representation of minority-class features in the parameter space makes their critical channels vulnerable to excessive removal. This can lead to an irreversible loss of important discriminative features. Such structural damage significantly hinders the recovery of these features during fine-tuning, ultimately reducing the model's ability to generalize for minority classes. To address this issue, we adjust the importance weighting of channels in equation~\ref{eq: pruning} as follows:
\begin{equation}
s_k^{(l)} = \sum_{j=1}^K w_j \cdot s_{k,j}^{(l)}\ ,
\end{equation}
where $ w_j $ represents class-aware weights, reflecting the relative contribution of class $ j $ in channel importance evaluation:
\begin{equation}
w_j = \frac{p_j}{\sum_{k=1}^K p_k}\ .
\end{equation}
Here, $p_j$ represents the relative frequency of class $j$ in the dataset, and normalization ensures that the sum of the weights equals $ 1 $. Minority classes are assigned higher weights through this formula, thus preventing the pruning strategy from excessively removing filters related to the minority.

During knowledge distillation, we compute the distillation losses for both the original few-sample dataset and the auxiliary dataset to prevent the model from overfitting to the OOD data after adding OOD data to the distillation process: 
\begin{equation}
\label{eq: knowledgedistillation}
L = \lambda \cdot L_{\text{KD,Few}} + (1-\lambda) \cdot L_{\text{KD,OOD}}\ ,
\end{equation}
where $ \lambda $ is a balancing coefficient that adjusts the relative contributions of the original dataset and the auxiliary dataset to the overall distillation process. It enables the student model to learn the knowledge related to minority classes more effectively.

\paragraph{Fine-tuning Stage:} After obtaining the compressed student model $ s $, fine-tuning is performed using the training samples. However, the imbalanced nature of the training dataset may degrade performance. To mitigate this, we incorporate OOD data to rebalance the fine-tuning process. Specifically, we merge the OOD auxiliary dataset $ D_{\text{aux}} $, which has been assigned labels during the compression phase, with the original imbalanced training dataset $ D_{\text{imb}} $ to construct the fine-tuning dataset $ D_{\text{train}} $.

Directly applying standard cross-entropy loss in this setting may result in overfitting to the auxiliary dataset, leading to convergence issues. To address this, we introduce a regularization term for the auxiliary dataset:
\begin{equation}
\mathcal{L}_{\text{reg}} = \mathbb{E}_{\tilde{x} \sim P_{\text{aux}}(X)} \left[ \gamma_{\tilde{y}} \cdot \ell \left( f(\tilde{x}; \theta), \tilde{y} \right) \right]\ ,
\end{equation}
where $ \tilde{y} $ is sampled from the complementary distribution. To further alleviate overfitting and preserve the performance on majority classes, we add a class-dependent weight factor $ \gamma_{\tilde{y}} $ to the regularization term and the overall loss function is:
\begin{align}
\label{eq: totalloss}
\mathcal{L}_{\text{total}} &= \mathbb{E}_{(x, y) \sim P_S(X, Y)} \left[ \ell \left( f(x; \theta), y \right) \right] \notag\\
\quad &+ \eta \cdot \mathbb{E}_{\tilde{x} \sim P_{\text{aux}}(X)} \left[ \gamma_{\tilde{y}} \cdot \ell \left( f(\tilde{x}; \theta), \tilde{y} \right) \right]\ . 
\end{align}
Here, $ \eta $ is a weighting factor balancing the contributions of the original dataset and the auxiliary dataset.

To address the overfitting tendency inherent in few-shot scenarios, we employ an early stopping mechanism during training. This ensures the final compact student model achieves optimal performance without sacrificing generalization capability.

\section{Experiment}

To address the following research questions, we conducted extensive experiments on three publicly available datasets:  
\begin{itemize}
    \item \textbf{RQ1:} Is class imbalance a significant issue for few-sample model compression methods?  
    \item \textbf{RQ2:} Can our framework effectively mitigate the issue of class imbalance in few-sample model compression, and is it compatible with current state-of-the-art few-sample model compression methods?  
    \item \textbf{RQ3:} How effective is each component of our framework?  
\end{itemize}

\subsection{Experiment Setting}
\noindent \textbf{Data Description:}
To validate that our framework effectively alleviates the class imbalance issue in few-sample model compression, we select image classification datasets CIFAR-10/100 and ILSVRC-2012. In the few-shot setting, we sample 10/30/50 data points from the original datasets. For each class, the number of samples is set according to the imbalance standard in long-tailed CIFAR-10/100~\citep{krizhevsky2009learning}. The samples are placed under the most common imbalance type—a long-tail distribution, with the imbalance ratio set to 100.

\noindent \textbf{Compared Method:}
We demonstrate that our framework effectively alleviates the class imbalance issue in few-sample model compression by integrating it with the following methods: 
\begin{enumerate}
    \item \textbf{CD}~\citep{bai2020few}: CD interweaves the hidden layers of the teacher and student networks.
    \item \textbf{FSKD}~\citep{li2020few}: FSKD adds a 1×1 convolutional layer at the end of each block in the student network, aligning the block-level outputs of the student and teacher models by estimating the parameters of the added layer.
    \item \textbf{MIR}~\citep{wang2022mir}: MiR encourages the pruned model to output the same features as the teacher model in the penultimate layer and replaces all layers in the teacher model before the penultimate layer (except for the head) with the trained student model.
    \item \textbf{PRACTISE}~\citep{wang2023practical}: PRACTISE replaces filter pruning with block dropping, prioritizing the removal of blocks based on their recoverability.
\end{enumerate}

\noindent \textbf{Evaluation Metric:}
Following previous few-sample model compression algorithms, we evaluate the effectiveness of our framework using top-1 accuracy. For each method, we conduct five independent experiments and report the mean deviation.

\noindent \textbf{Parameter Setting:}
In our experimental setup, the auxiliary dataset was configured to match the size of the few-sample dataset, with a batch size of 128 and a learning rate of 0.0005. For all compared methods, we maintained their original hyperparameters and pruning ratios as reported in their respective reference implementations to prove the generalization of our framework. All experiments are conducted on an RTX-4090 24GB GPU and an I9-13900KF CPU.

\subsection{Validation and Analysis of the Problem (RQ1)}
The experiments in this section aim to analyze the resistance of existing few-sample model compression algorithms to class imbalance, rather than comparing the performance of different few-sample model compression methods. Therefore, we did not apply a unified pruning strategy for all algorithms but instead tested them using the pruning rates and parameters provided in their original papers. To streamline the setup, we standardized the training set to CIFAR-10 and used a pre-trained teacher model, VGG-16. First, we set the training set to CIFAR-10. For the balance configuration, we retained the N-way, K-shot setup to ensure equal sample counts for each class. In the imbalance configuration, we adjusted the sample numbers per class according to an imbalanced distribution to reflect the most common long-tail distribution. In other words, we compared the performance of previous few-sample model compression algorithms without any anti-imbalance measures under balanced and imbalanced datasets. The results in Table~\ref{tab:validation} show that when the dataset suffers from class imbalance, the performance of existing few-sample model compression methods significantly decreases. Among them, the PRACTISE method is less affected, likely because the block-dropping strategy is less sensitive to imbalance. However, under imbalance conditions, its latency increases significantly, and since "latency-accuracy" is the key evaluation metric emphasized by PRACTISE, its performance is also affected by the imbalance issue in some ways.

\begin{table}[t]
\centering
\caption{Test accuracy (\%) of VGG-16 on long-tailed CIFAR-10 under balanced and imbalanced conditions. $\downarrow$ indicates the accuracy loss in the imbalanced case.}
\label{tab:validation}
\resizebox{\linewidth}{!}{
\begin{tabular}{c|cc|cc|cc|cc}
\toprule
\multicolumn{1}{c|}{Num} & \multicolumn{2}{c|}{10}  & \multicolumn{2}{c|}{20} & \multicolumn{2}{c|}{50} & \multicolumn{2}{c}{100} \\

\cmidrule(){1-1} \cmidrule(){2-3} \cmidrule(){4-5} \cmidrule(){6-7} \cmidrule(){8-9}
 Imbalance Type & \multicolumn{1}{c}{Balance} & \multicolumn{1}{c|}{Imbalance} & 
 \multicolumn{1}{c}{Balance} & \multicolumn{1}{c|}{Imbalance} &
 \multicolumn{1}{c}{Balance} & \multicolumn{1}{c|}{Imbalance} &
 \multicolumn{1}{c}{Balance} & \multicolumn{1}{c}{Imbalance}\\
\midrule
CD               & 73.89   & 70.34\textbf{(3.55$\downarrow$)}  &  81.65 & 76.39\textbf{(5.26$\downarrow$)} & 82.86 & 80.47\textbf{(2.39$\downarrow$)} & 83.56 & 82.24\textbf{(1.32$\downarrow$)} \\

FSKD   & 76.18   & 73.51\textbf{(2.67$\downarrow$)}  &  84.20 & 81.66\textbf{(2.54$\downarrow$)} & 88.63 & 86.31\textbf{(2.32$\downarrow$)} & 89.18 & 87.41\textbf{(1.77$\downarrow$)}             \\

MiR    &75.42 & 72.84\textbf{(2.58$\downarrow$)}
 & 79.18 & 76.57\textbf{(2.61$\downarrow$)} & 84.27 & 82.49\textbf{(1.78$\downarrow$)} & 85.71 & 83.85\textbf{(1.86$\downarrow$)}               \\

PRACTISE & 76.91 & 75.78\textbf{(1.13$\downarrow$)} & 82.07 & 81.26\textbf{(0.81$\downarrow$)} & 86.35 & 85.98\textbf{(0.37$\downarrow$)} & 89.71 & 89.35\textbf{(0.35$\downarrow$)}                    \\

\bottomrule
\end{tabular}
}
\end{table}

\subsection{Effectiveness and Generalizability (RQ2)}
In this experiment, we evaluated the effectiveness of the OE-FSMC method on different model architectures and datasets, specifically including VGG-16 on long-tail CIFAR-100, ResNet-32 on long-tail CIFAR-10, and ResNet-34 on long-tail ILSVRC-2012. Table~\ref{tab:mainresults} presents the performance of several mainstream few-sample model compression strategies before and after combining with OE-FSMC, with a focus on addressing the class imbalance problem. The results clearly show that after incorporating the OE-FSMC method, the accuracy of each few-sample model compression strategy improved significantly. Notably, the performance improvement brought by OE-FSMC is more pronounced when the number of training samples is smaller, suggesting that OE-FSMC is particularly effective in mitigating class imbalance in scenarios with limited data.

It is important to note that we retained the pruning rate, learning rate, and other hyperparameter settings used in the original few-sample model compression strategies. We believe this approach accurately reflects the model’s performance in the context of class imbalance, while also demonstrating the effectiveness and generalizability of the OE-FSMC method. Furthermore, although the training set sizes of 1000, 2000, and 3000 samples might seem large, considering that the ILSVRC-2012 dataset consists of 1000 classes, we chose these settings to ensure that each class has 1, 2, and 3 samples in the balanced case. This setup explains why the effect of OE-FSMC on the ILSVRC-2012 dataset is less pronounced compared to other datasets, but overall, OE-FSMC still shows promising results.

\begin{table}[t]
\centering
\caption{Test accuracy (\%) of VGG-16 on long-tailed CIFAR-10, ResNet-32 on CIFAR-100 and ResNet-34 on long-tailed ILSVRC-2012 with different size of trainset.}
\label{tab:mainresults}
\resizebox{\linewidth}{!}{
\begin{tabular}{c|ccc|ccc|ccc}
\toprule
\multicolumn{1}{c|}{Dataset} 

& \multicolumn{3}{c|}{Long-tailed CIFAR-10} 
& \multicolumn{3}{c|}{Long-tailed CIFAR-100} 
& \multicolumn{3}{c}{Long-tailed ILSVRC-2012}\\
\cmidrule{1-1}
\cmidrule{2-4}  
\cmidrule{5-7} 
\cmidrule{8-10}

\multicolumn{1}{c|}{Num} 
& \multicolumn{1}{c}{10} & \multicolumn{1}{c}{20} & \multicolumn{1}{c|}{50} 
& \multicolumn{1}{c}{100} & \multicolumn{1}{c}{200} & \multicolumn{1}{c|}{500} 
& \multicolumn{1}{c}{1000} & \multicolumn{1}{c}{2000} & \multicolumn{1}{c}{3000} \\
 
\midrule
CD               
& 70.34  & 76.39  & 80.47
& 53.38  & 62.71  & 67.40
& 69.49  & 70.07  & 70.14
\\
\textbf{+ Ours}
& \textbf{78.90} & \textbf{81.02} & \textbf{82.80}
& \textbf{56.33} & \textbf{65.18} & \textbf{68.84}
& \textbf{69.84} & \textbf{70.56} & \textbf{70.58}
\\
FSKD 
& 73.51 & 81.66 & 86.31 
& 57.78 & 63.29 & 68.15
& 68.54 & 69.95 & 70.27
\\
\textbf{+ Ours}                
&\textbf{75.93} & \textbf{84.67} & \textbf{88.24}
& \textbf{59.46} & \textbf{65.88} & \textbf{69.17}
& \textbf{69.06} & \textbf{70.18} & \textbf{70.60}

\\
MiR                   
& 72.84 & 76.57 & 82.49
& 57.69 & 64.92 & 68.70
& 66.36 & 67.23 & 67.48

\\
\textbf{+ Ours}              
& \textbf{75.25} & \textbf{77.93} & \textbf{83.60}            
& \textbf{60.37} & \textbf{66.53} & \textbf{69.83}
& \textbf{66.85} & \textbf{67.97} & \textbf{68.12}
\\
PRACTISE                      
& 75.78 & 81.26 & 85.98
& 58.42 & 65.14 & 69.02
& 70.90 & 71.89 & 72.41

\\
\textbf{+ Ours}           
& \textbf{76.59} & \textbf{81.77} &\textbf{86.25}
& \textbf{69.57} & \textbf{66.08} &\textbf{69.62}
& \textbf{71.52} & \textbf{72.17} &\textbf{72.85}
\\
\bottomrule
\end{tabular}
}
\end{table}

\subsection{Ablation Study (RQ3)}
We evaluate the performance of the individual components of OE-FSMC by comparing with the following variants:
\begin{itemize}
    \item $\text{OE-FSMC}_{w/o\ Comp.}$: In this variant, we exclude the use of OOD data during the compression process. Additionally, we remove the class-dependent weight factor and revert the joint distillation loss to the standard distillation loss.
    \item $\text{OE-FSMC}_{w/o\ FT}$: In this variant, we no longer incorporate OOD data into the fine-tuning process, and the regularization loss $\mathcal{L}_{\text{reg}}$ is removed.
    \item $\text{OE-FSMC}_{w/o\ all}$: In this variant, all proposed modifications are removed to comprehensively demonstrate the performance of the original few-sample model compression method on an imbalanced dataset.
\end{itemize}

\begin{figure}[t]
	\centering
	\resizebox{\linewidth}{!}{ 
		\subfigure[\label{fig:ablation}]{\includegraphics[width=0.5\linewidth]{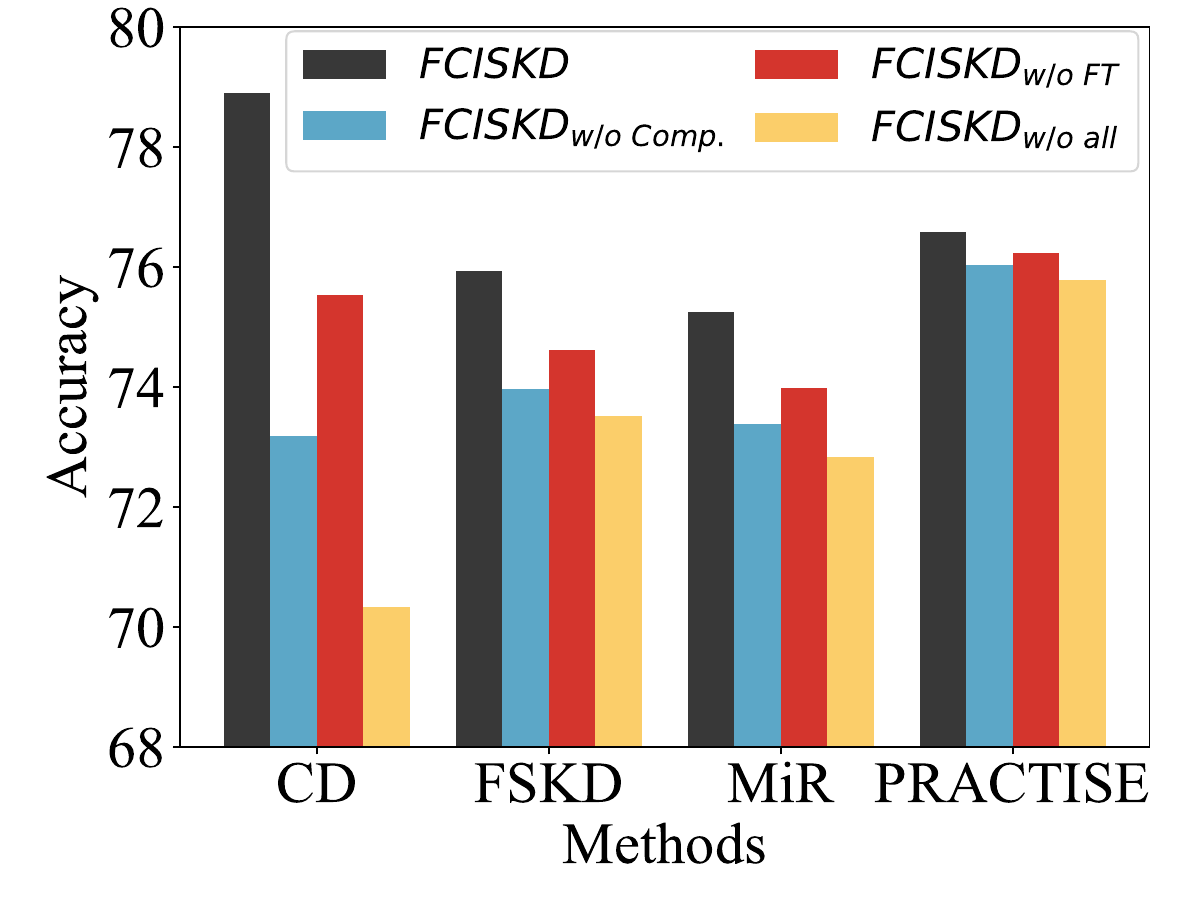}}\hfill
		\subfigure[\label{fig:auxsize} ]{\includegraphics[width=0.5\linewidth]{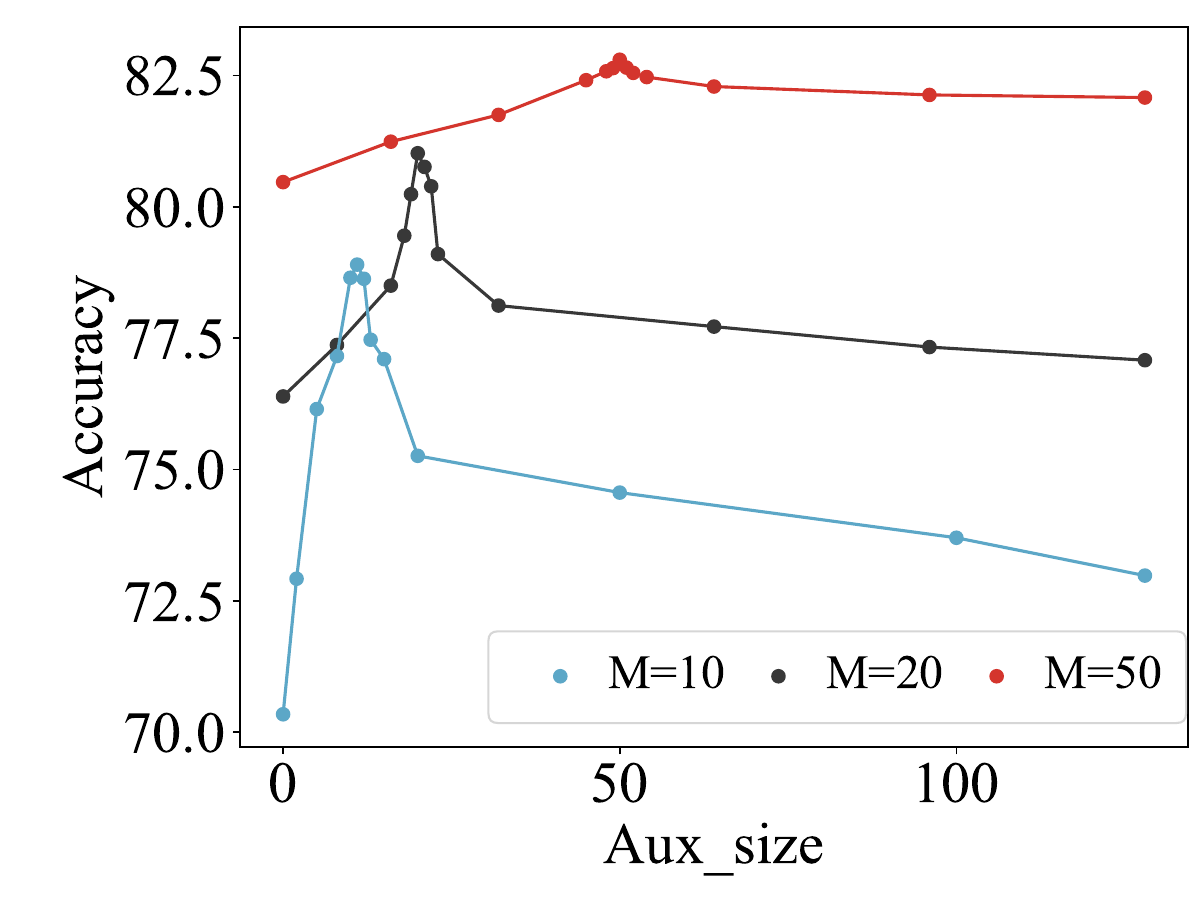}}
	}
    \vfill 
	\resizebox{\linewidth}{!}{ 
		\subfigure[\label{fig:lambda}]{\includegraphics[width=0.5\linewidth]{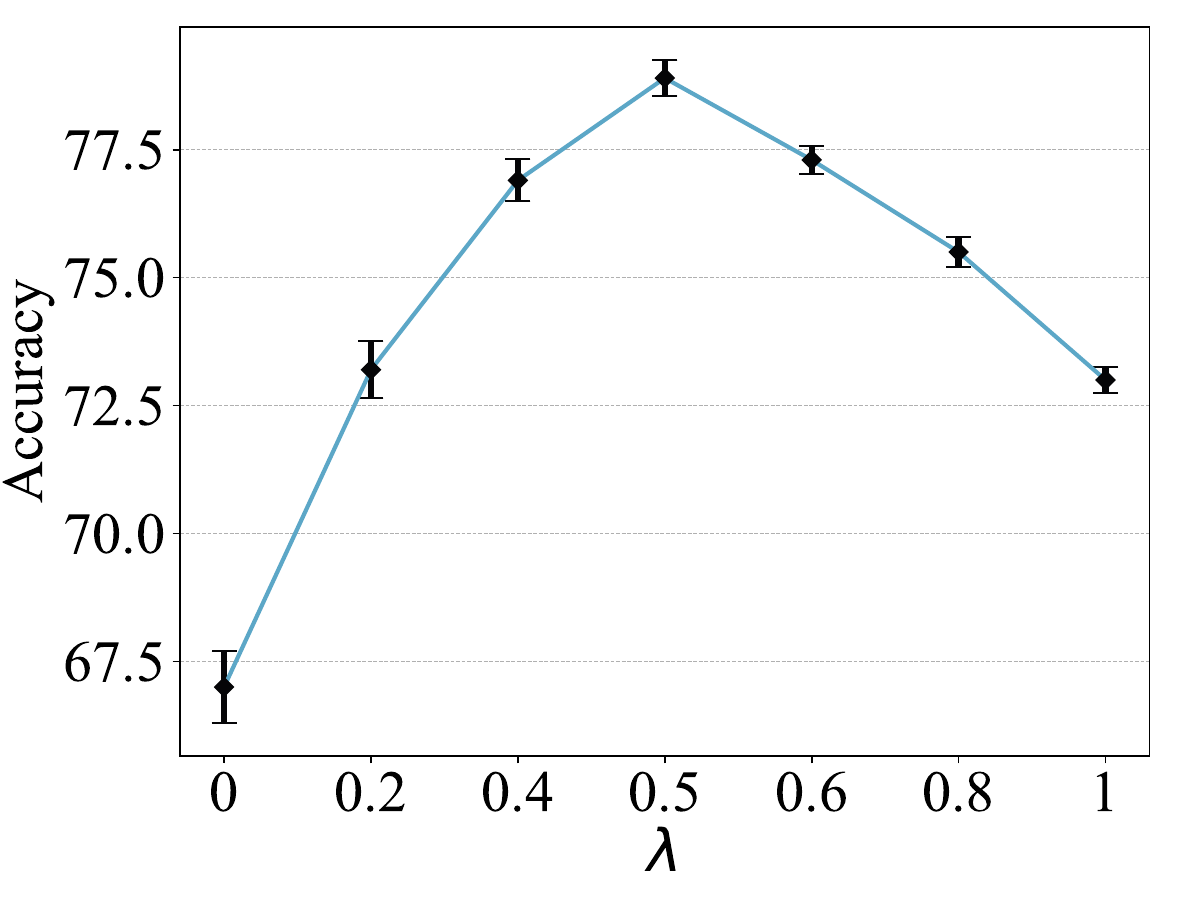}}\hfill
		\subfigure[\label{fig:eta}]{\includegraphics[width=0.5\linewidth]{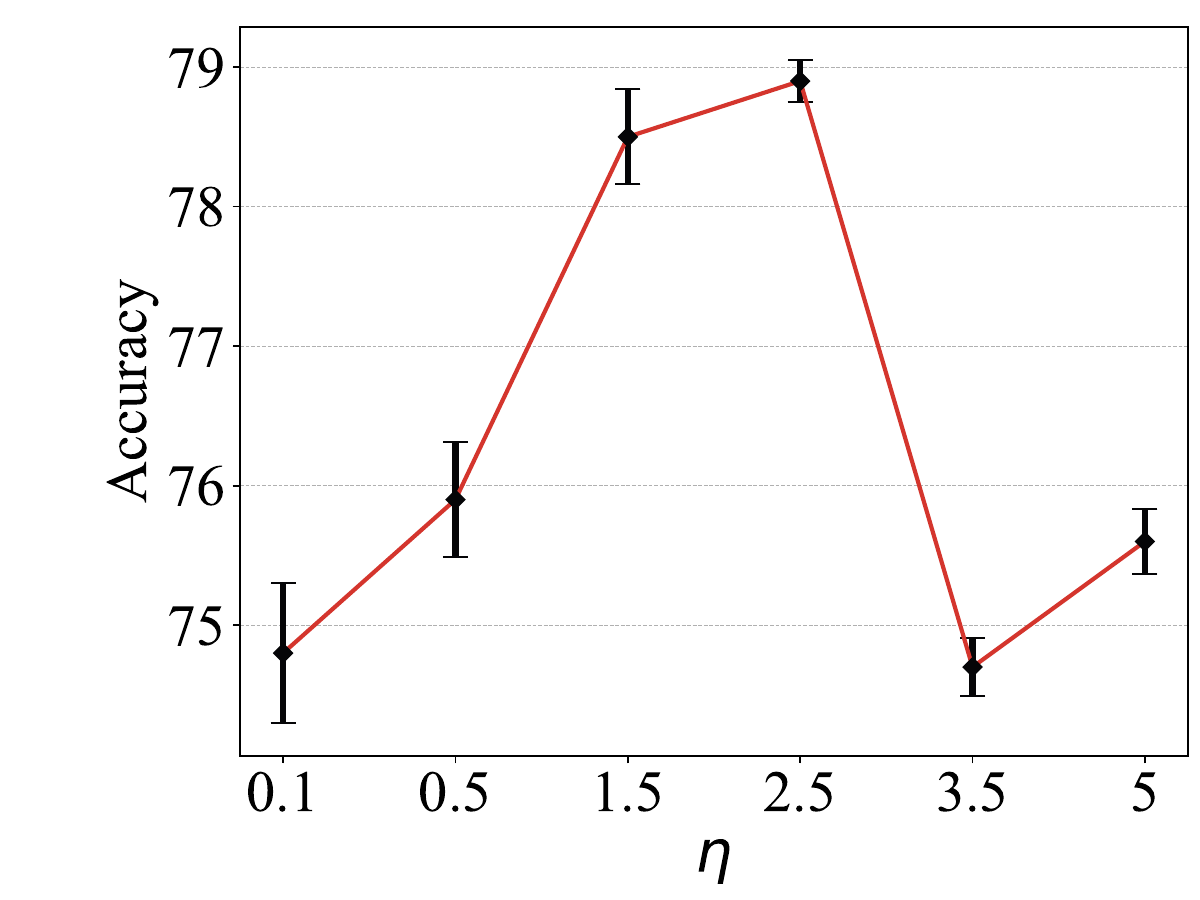}}
	}
\caption{Results of extra experiments.In this figure: a) Results of ablation study. b) Effect of different sizes of the auxiliary set. c) Sensitivity analysis of $\lambda \in [0, 1]$. d) Sensitivity analysis of $\eta \in [0.1, 5]$. }
\end{figure}

In each experiment, we progressively remove specific components of OE-FSMC and test them on the same network and dataset. The experimental results are presented in Figure~\ref{fig:ablation}. From these results, we make the following observations: the two components of our method are both highly effective in mitigating the detrimental effects of class imbalance in the few-sample compression process. Using either component alone improves the performance of the compressed model. Notably, when both components are used together, the performance is further enhanced; omitting either one leads to a decrease in accuracy. The compression component leverages OOD data to balance the model compression process, enabling the compressed model to better preserve the feature representations of minority classes. Once these features are lost, they are challenging to recover through fine-tuning. This may explain why the performance improvement of $\text{OE-FSMC}_{w/o\ FT}$ surpasses that of $\text{OE-FSMC}_{w/o\ Comp.}$. The fine-tuning component incorporates OOD data to balance the class priors, mitigating overfitting to the majority class during training and further enhancing the performance on minority classes.

\subsection{Hyperparameter Sensitivity Analysis}
To investigate the impact of the size of $ D_{\text{aux}}$ on model performance, we conducted experiments under three different training dataset sizes, the results in Figure~\ref{fig:auxsize} demonstrate that the model achieves optimal performance when the size of the auxiliary dataset is comparable to the original training dataset. This suggests that an appropriately sized OOD dataset can effectively mitigate the class imbalance problem in the original dataset during few-sample model compression.

Figure~\ref{fig:lambda} illustrates the effect of different values of $\lambda$ in Equation~\ref{eq: knowledgedistillation} on model accuracy. We vary its value in \{0, 0.2, 0.4, 0.5, 0.6, 0.8, 1.0\}, and the result indicates that the model achieves the best performance when 
$\lambda=0.5$. This suggests that assigning equal weight to the original training data and the OOD data during knowledge distillation achieves a balanced outcome, effectively preserving the features of minority classes while avoiding over-reliance on OOD data. To further explore the effect of $\eta$ in Equation~\ref{eq: totalloss}, we search $\eta$ in \{0.1, 0.5, 1.5, 2.5, 3.5, 5\}. From Figure~\ref{fig:eta}, we observe that the model performance improves significantly when $\eta=2.5$. This finding suggests that moderate regularization effectively constrains the model’s parameter search space, thereby enhancing generalization. When $\eta$ is too large, the regularization term dominates, suppressing the model’s ability to learn features from both the training and OOD datasets, leading to a decline in performance. Overall, our ablation experiments provide additional evidence for the effectiveness of our proposed method, further demonstrating that it can achieve strong performance under reasonable parameter configurations.

\section{Conclusion}
The class imbalance problem in the few-sample model compression is a critical problem that has been overlooked by previous methods. In this paper, we propose a novel exploratory framework called OOD-Enhanced Few-Sample Model Compression (OE-FSMC), aimed at mitigating the negative impact of class imbalance on few-sample model compression. Specifically, we leverage easily accessible OOD data to balance both the compression and fine-tuning processes. In addition, we implement a joint distillation loss and a regularization term to prevent the model from overfitting to the OOD data. Extensive experiments conducted across a wide range of few-sample model compression strategies and benchmark datasets demonstrate the effectiveness and generality of our framework. In future work, we will explore further applications, including its generalization to compression strategies that are not commonly utilized in current few-sample model compression methods, such as model quantization.


\acks{This work was supported by the National Natural Science Foundation of China (62306104), Hong Kong Scholars Program (XJ2024010), Jiangsu Science Foundation (BK20230949), China Postdoctoral Science Foundation (2023TQ0104), Jiangsu Excellent Postdoctoral Program (2023ZB140).}


\bibliography{main}

\begin{thebibliography}{26}
\providecommand{\natexlab}[1]{#1}
\providecommand{\url}[1]{\texttt{#1}}
\expandafter\ifx\csname urlstyle\endcsname\relax
  \providecommand{\doi}[1]{doi: #1}\else
  \providecommand{\doi}{doi: \begingroup \urlstyle{rm}\Url}\fi

\bibitem[Abdi and Hashemi(2015)]{abdi2015combat}
L.~Abdi and S.~Hashemi.
\newblock To combat multi-class imbalanced problems by means of over-sampling techniques.
\newblock \emph{IEEE Transactions on Knowledge and Data Engineering}, 28\penalty0 (1):\penalty0 238--251, 2015.

\bibitem[Bai et~al.(2020)Bai, Wu, King, and Lyu]{bai2020few}
H.~Bai, J.~Wu, I.~King, and M.~Lyu.
\newblock Few shot network compression via cross distillation.
\newblock In \emph{Proceedings of the 34th AAAI Conference on Artificial Intelligence}, pages 3203--3210, 2020.

\bibitem[Chawla et~al.(2002)Chawla, Bowyer, Hall, and Kegelmeyer]{chawla2002smote}
N.~V. Chawla, K.~W. Bowyer, L.~O. Hall, and W.~P. Kegelmeyer.
\newblock Smote: {S}ynthetic minority over-sampling technique.
\newblock \emph{Journal of Artificial Intelligence Research}, 16:\penalty0 321--357, 2002.

\bibitem[Chawla et~al.(2003)Chawla, Lazarevic, Hall, and Bowyer]{chawla2003smoteboost}
N.~V. Chawla, A.~Lazarevic, L.~O. Hall, and K.~W. Bowyer.
\newblock Smoteboost: Improving prediction of the minority class in boosting.
\newblock In \emph{Proceedings of 7th European Conference on Principles and Practice of Knowledge Discovery in Databases}, pages 107--119, 2003.

\bibitem[Dong et~al.(2017)Dong, Chen, and Pan]{dong2017learning}
X.~Dong, S.~Chen, and S.~Pan.
\newblock Learning to prune deep neural networks via layer-wise optimal brain surgeon.
\newblock \emph{Advances in Neural Information Processing Systems}, 30, 2017.

\bibitem[Fern{\'a}ndez et~al.(2018)Fern{\'a}ndez, Garc{\'\i}a, Galar, Prati, Krawczyk, Herrera, Fern{\'a}ndez, Garc{\'\i}a, Galar, Prati, et~al.]{fernandez2018cost}
A.~Fern{\'a}ndez, S.~Garc{\'\i}a, M.~Galar, R.~C. Prati, B.~Krawczyk, F.~Herrera, A.~Fern{\'a}ndez, S.~Garc{\'\i}a, M.~Galar, R.~C. Prati, et~al.
\newblock Cost-sensitive learning.
\newblock In \emph{Learning from Imbalanced Data Sets}, pages 63--78. Springer, Cham, 2018.

\bibitem[Galar et~al.(2011)Galar, Fernandez, Barrenechea, Bustince, and Herrera]{galar2011review}
M.~Galar, A.~Fernandez, E.~Barrenechea, H.~Bustince, and F.~Herrera.
\newblock A review on ensembles for the class imbalance problem: bagging-, boosting-, and hybrid-based approaches.
\newblock \emph{IEEE Transactions on Systems, Man, and Cybernetics, Part C}, 42\penalty0 (4):\penalty0 463--484, 2011.

\bibitem[He et~al.(2017)He, Zhang, and Sun]{yihuiChannel2017}
Y.~He, X.~Zhang, and J.~Sun.
\newblock Channel pruning for accelerating very deep neural networks.
\newblock In \emph{Proceedings of the 16th the IEEE International Conference on Computer Vision}, pages 1398--1406, 2017.

\bibitem[He et~al.(2024)He, Liu, Lyu, Qian, and Zhou]{he2024multi}
Y.-X. He, D.-X. Liu, S.-H. Lyu, C.~Qian, and Z.-H. Zhou.
\newblock Multi-class imbalance problem: {A} multi-objective solution.
\newblock \emph{Information Sciences}, 680:\penalty0 121156, 2024.

\bibitem[Hinton(2015)]{hinton2015distilling}
G.~Hinton.
\newblock Distilling the knowledge in a neural network.
\newblock \emph{arXiv preprint arXiv:1503.02531}, 2015.

\bibitem[Japkowicz and Stephen(2002)]{japkowicz2002class}
N.~Japkowicz and S.~Stephen.
\newblock The class imbalance problem: A systematic study.
\newblock \emph{Intelligent Data Analysis}, 6\penalty0 (5):\penalty0 429--449, 2002.

\bibitem[Krizhevsky and Hinton(2009)]{krizhevsky2009learning}
A.~Krizhevsky and G.~Hinton.
\newblock Learning multiple layers of features from tiny images.
\newblock Technical Report UT-ML-TR-2009-001, University of Toronto, 2009.

\bibitem[Li et~al.(2017)Li, Kadav, Durdanovic, Samet, and Graf]{hao2017pruning}
H.~Li, A.~Kadav, I.~Durdanovic, H.~Samet, and H.~P. Graf.
\newblock Pruning filters for efficient convnets.
\newblock In \emph{Proceedings of the 5th International Conference on Learning Representations}, 2017.

\bibitem[Li et~al.(2020)Li, Li, Liu, and Zhang]{li2020few}
T.~Li, J.~Li, Z.~Liu, and C.~Zhang.
\newblock Few sample knowledge distillation for efficient network compression.
\newblock In \emph{Proceedings of the 32nd IEEE/CVF Conference on Computer Vision and Pattern Recognition}, pages 14639--14647, 2020.

\bibitem[Lin et~al.(2017)Lin, Tsai, Hu, and Jhang]{lin2017clustering}
W.-C. Lin, C.-F. Tsai, Y.-H. Hu, and J.-S. Jhang.
\newblock Clustering-based undersampling in class-imbalanced data.
\newblock \emph{Information Sciences}, 409:\penalty0 17--26, 2017.

\bibitem[Liu et~al.(2008)Liu, Wu, and Zhou]{liu2008exploratory}
X.-Y. Liu, J.~Wu, and Z.-H. Zhou.
\newblock Exploratory undersampling for class-imbalance learning.
\newblock \emph{IEEE Transactions on Systems, Man, and Cybernetics, Part B}, 39\penalty0 (2):\penalty0 539--550, 2008.

\bibitem[Mohammed et~al.(2020)Mohammed, Rawashdeh, and Abdullah]{mohammed2020machine}
R.~Mohammed, J.~Rawashdeh, and M.~Abdullah.
\newblock Machine learning with oversampling and undersampling techniques: {O}verview study and experimental results.
\newblock In \emph{Proceedings of the 11th International Conference on Information and Communication Systems}, pages 243--248, 2020.

\bibitem[Nagel et~al.(2019)Nagel, Baalen, Blankevoort, and Welling]{nagel2019data}
M.~Nagel, M.~v. Baalen, T.~Blankevoort, and M.~Welling.
\newblock Data-free quantization through weight equalization and bias correction.
\newblock In \emph{Proceedings of the 17th IEEE/CVF International Conference on Computer Vision}, pages 1325--1334, 2019.

\bibitem[Ochal et~al.(2023)Ochal, Patacchiola, Vazquez, Storkey, and Wang]{ochal2023few}
M.~Ochal, M.~Patacchiola, J.~Vazquez, A.~Storkey, and S.~Wang.
\newblock Few-shot learning with class imbalance.
\newblock \emph{IEEE Transactions on Artificial Intelligence}, 2023.

\bibitem[Romero et~al.(2015)Romero, Ballas, Kahou, Chassang, Gatta, and Bengio]{adrianafitnet2015}
A.~Romero, N.~Ballas, S.~E. Kahou, A.~Chassang, C.~Gatta, and Y.~Bengio.
\newblock Fitnets: Hints for thin deep nets.
\newblock In \emph{Proceedings of 3rd International Conference on Learning Representations}, pages 1--13, 2015.

\bibitem[Sharma et~al.(2022)Sharma, Gosain, and Jain]{sharma2022review}
S.~Sharma, A.~Gosain, and S.~Jain.
\newblock A review of the oversampling techniques in class imbalance problem.
\newblock In \emph{Proceedings of the 5th International Conference on Innovative Computing and Communications}, pages 459--472, 2022.

\bibitem[Wang and Wu(2023)]{wang2023practical}
G.-H. Wang and J.~Wu.
\newblock Practical network acceleration with tiny sets.
\newblock In \emph{Proceedings of the 36th IEEE/CVF Conference on Computer Vision and Pattern Recognition}, pages 20331--20340, 2023.

\bibitem[Wang et~al.(2022)Wang, Liu, Ma, Yong, Chai, and Wu]{wang2022mir}
H.~Wang, J.~Liu, X.~Ma, Y.~Yong, Z.~Chai, and J.~Wu.
\newblock Compressing models with few samples: {M}imicking then replacing.
\newblock In \emph{Proceedings of the 35th IEEE/CVF Conference on Computer Vision and Pattern Recognition}, pages 701--710, 2022.

\bibitem[Wei et~al.(2022)Wei, Tao, Xie, Feng, and An]{wei2022open}
H.~Wei, L.~Tao, R.~Xie, L.~Feng, and B.~An.
\newblock Open-sampling: Exploring out-of-distribution data for re-balancing long-tailed datasets.
\newblock In \emph{Proceedings of the 39th International Conference on Machine Learning}, pages 23615--23630, 2022.

\bibitem[Zhou et~al.(2020)Zhou, Cui, Wei, and Chen]{zhou2020bbn}
B.~Zhou, Q.~Cui, X.-S. Wei, and Z.-M. Chen.
\newblock Bbn: Bilateral-branch network with cumulative learning for long-tailed visual recognition.
\newblock In \emph{Proceedings of the 32rd IEEE/CVF Conference on Computer Vision and Pattern Recognition}, pages 9719--9728, 2020.

\bibitem[Zhou and Liu(2005)]{zhou2005training}
Z.-H. Zhou and X.-Y. Liu.
\newblock Training cost-sensitive neural networks with methods addressing the class imbalance problem.
\newblock \emph{IEEE Transactions on Knowledge and Data Engineering}, 18\penalty0 (1):\penalty0 63--77, 2005.

\end{thebibliography}

\end{document}